# SALIENCY DETECTION WITH FULLY CONVOLUTIONAL NEURAL NETWORK

[1]HOOMAN MISAGHI, [2]REZA ASKARI MOGHADAM, [3]ALI MAHMOUDI, [4]KUROSH MADANI

[1,2,3]Faculty of New Sciences and Technologies University of Tehran Tehran, Iran
[4]Laboratoire Images, Signaux et Syst`emes Intelligents Universit´e Paris-Est,Paris, France
E-mail: [1]hooman.misaghi@ut.ac.ir, [2]r.askari@ut.ac.ir, [3]ali.mahmoudii@ut.ac.ir, [4]madani@u-pec.fr

**Abstract** - Saliency detection is an important task in image processing as it can solve many problems and it usually is the first step in for other processes. Convolutional neural networks have been proved to be very effective on several image processing tasks such as classification, segmentation, semantic colorization and object manipulation. Besides, using the weights of a pretrained networks is a common practice for enhancing the accuracy of a network. In this paper a fully convolutional neural network which uses a part of VGG-16 is proposed for saliency detection in images.

## I. INTRODUCTION

Saliency detection, as its name suggests, is a task in which the most salient part of an image is specified. It is worth noting that the most salient part of an image is the region of that image which is noticed immediately when a random viewer gazes upon that image. This definition shows that saliency detection is a very important task, since it is the first step in other more complex tasks, such as unsupervised learning, semantic colorization, content aware resizing, and object manipulation.

Therefore, Saliency detection – and consequently, the systems mentioned above – is an important tool in building the next generation of robots that act intelligently and could interact with humans in an elegant fashion. During the past few years several methods have been proposed for saliency detection. Achanta et-al. proposed a method based on computing uniqueness of an object in the image in pixel level by comparing it with its surrounding [1].
In addition, innovative methods, such as graph-based segmentation and mean-shift, compute saliency map after segmentation of the mage [2] [3]. Another method extracts salient objects by ensemble of global and local saliency maps [4]. Another prominent method in processing images is Convolutional Neural Networks (CNN). CNN's ability in extracting image's features in different levels has been put to use in many tasks [5]. A convolutional neural network has been proposed which consists of two complementary components, a pixel level fully convolutional stream and a segment-wise spatial pooling stream [6]. There is also another method using image segmentation and then processing by CNN [7] and another method which does not use any pre trained networks [8].

In this research, we put our efforts into designing a system for saliency detection task based on deep learning and specifically convolutional neural networks. Three networks are presented in this paper. A pixel level saliency detection network built on top of VGG-16 is presented. There have been several methods using fully convolutional neural networks for saliency detection task but ours stands out such as [9] [7] which use VGGNet as well but ours stands out in the simplicity of training and accuracy. By tuning the VGG architecture to be better suited for saliency detection task in terms of passing data.

The rest of this paper is organized as follows: Section II discusses the network's architectures, Section III gives information about the training procedure, Section IV demonstrates the performance results of both networks, and finally, Section V discusses the obtained results and draws conclusion.

## II. NETWORK ARCHITECTURE

Using pre-trained networks have been useful for different tasks. It is a common practice to use pre-trained networks and retraining the last layers for new tasks or adding some layers on top of a pre-trained network for different tasks. Using a pre-trained network on saliency detection task has been shown effective. By using pre-trained convolutional layers, the network will have the ability to extract different features from the images since the start of training. VGG-16 has been trained originally on image-net dataset for image classification task [10]. It consists of several convolutional layers followed by max pooling layers. At the end of the of the VGG networks there is three fully connected layers to classify the image in 1000 classes. The tableI illustrates VGG architecture.

The network used in this uses the convolutional layer of D configuration. The following network architecture is built on top of VGG- 16 pre-trained by TensorFlow team. VGG-16 convolutional layers have been used as down sampling layers of the whole system. Input image goes through the VGG layers and turns to a feature map with 1/32 of its original size with 512 depth.

## TABLE I: VGG-16 ARCHITECTURE [10]

| A | A-LRN | B | C | D | E |
|---|---|---|---|---|---|
| 11 weight layers | 11 weight layers | 13 weight layers | 16 weight layers | 16 weight layers | 19 weight layers |
| input (224 × 224 RGB image) | | | | | |
| conv3-64 | conv3-64 LRN | conv3-64 conv3-64 | conv3-64 conv3-64 | conv3-64 conv3-64 | conv3-64 conv3-64 |
| maxpool | | | | | |
| conv3-128 | conv3-128 | conv3-128 conv3-128 | conv3-128 conv3-128 | conv3-128 conv3-128 | conv3-128 conv3-128 |
| maxpool | | | | | |
| conv3-256 conv3-256 | conv3-256 conv3-256 | conv3-256 conv3-256 | conv3-256 conv3-256 conv1-256 | conv3-256 conv3-256 conv3-256 | conv3-256 conv3-256 conv3-256 conv3-256 |
| maxpool | | | | | |
| conv3-512 conv3-512 | conv3-512 conv3-512 | conv3-512 conv3-512 | conv3-512 conv3-512 conv1-512 | conv3-512 conv3-512 conv3-512 | conv3-512 conv3-512 conv3-512 conv3-512 |
| maxpool | | | | | |
| conv3-512 conv3-512 | conv3-512 conv3-512 | conv3-512 conv3-512 | conv3-512 conv3-512 conv1-512 | conv3-512 conv3-512 conv3-512 | conv3-512 conv3-512 conv3-512 conv3-512 |
| maxpool | | | | | |
| FC-4096 | | | | | |
| FC-4096 | | | | | |
| FC-1000 | | | | | |
| softmax | | | | | |

Then with additional layers this feature map gets processed and up sampled to original size. Up sampling is done using 5 transpose convolution layers with 2 by 2 filters and stride 2 followed by a 1 by 1 convolutional layer with stride 1 which has a sigmoid activation function the output of the last layer is in the interval of (0,1). This output can easily be mapped as a grayscale image. Since with max-pooling layers some information would be lost, by using average pooling instead of max pooling results of the network improve. Again, in this network just like the previous network the last layer has sigmoid ctivation. No shoertcut have been used in this architecture. instead to pass more data to the last layers, aveeage pooling is used instead of maxpooling in VGG network. Firs of all average pooling asses the effect of whole data in a window while max pooling looses non max elements. Second, shortcuts usually concatente teh feature map from earlier layers to latter layers of network with the same size and mixing the feature maps are usually don by concatenation, this leads to having more operations to do because of the depth of the feature map that are the input of latter convolutions. This affects the speed of the network.

### III. TRAINING PROCEDURE

Training the network in this research is supervised which needs the labeled data. There are several datasets prepared for saliency detection such as MSRA-10k [11] and HKU-IS [12]. in this paper HKU-IS is used for training. HKU-IS consists of 4447 images with pixel tags as salient or non-salient in the image. HKU-IS most important feature is that it has several salient object in each image. Update algorithm is Adam which is available in Tensorflow [13]. This method uses an individual coefficient to update each weight and uses the first and second momentum to enhance the learning process [13]. The loss function used for training is

$$loss = \frac{1}{width \cdot height} \sum_{x=0}^{width} \sum_{y=0}^{height} |prediction(x,y) - GT(x,y)| \quad (1)$$

Which prediction is the output of the network and GT is the ground truth provided in the dataset. x; y illustrate the position of the pixel. The absolute function produces better results in comparison with l2 loss in this configuration. Absolute function used in the loss function does not have the derivative in 0 but since the last layer in the network has sigmoid activations the results will never be exact 0. Training was done for 20 epochs with batch size of 20 on HKU-IS dataset. VGG layers were freezed and no update operation was applied on them.

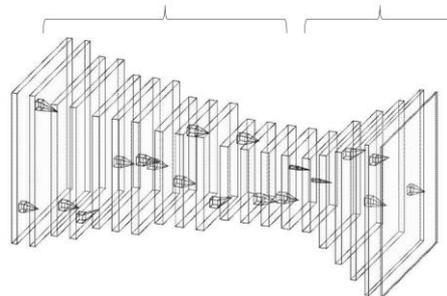

**Fig. 1. Proposed Network's architecture**

### IV. RESULTS AND EVALUATION

In this network an FCNN(Fully Convolutional Neural Network) architecture based on VGG 16 has been proposed. The results of the trained network are presented in this section. Evaluation results are presented as average mean absolute error, average F-measure, average precision and recall as well as the precision-recall curve.

The network was tested on MSRA-10K dataset that contains pixel level accurate salient object labeling for 10000 images from MSRA-10K dataset [14]. Four evaluation parameters used to evaluate the performance of the proposed network. These evaluation parameters are considered as standard evaluation methods for saliency detection in pixel level. Most of these evaluation measures require the map to be binary format. The network output needs to be processed with binarization techniques such as simple thresholding which is used in this paper. Every element in absolute-value bars represents a none zero entry in the saliency map. precision and recall can be calculated by (2) and (3)

$$precision = \frac{|M \cap G|}{|M|} \quad (2)$$

$$recall = \frac{|M \cap G|}{|G|} \quad (3)$$

In (2) and (3), M represents the binary saliency map and G is the binary ground truth map which usually is provided in the dataset. Considering that precision and recall are not the best methods for evaluating saliency detection methods due to their shortcomings. One of them only values positive samples and the other focuses on negative. Thus other methods are usually used to evaluate saliency detection systems which are more meaningful. F-measure and Mean absolute error (MAE) can becalculated by (4) and (5), these methods are more meaningful for evaluating saliency detection systems.

$$F_\beta = \frac{(1+\beta^2) precision \times recall}{\beta^2 precision \times recall} \quad (4)$$

$$MAE = \frac{1}{WH} \sum_{x=1}^{W} \sum_{y=1}^{H} \|S(x,y) - G(x,y)\| \quad (5)$$

In (4) it is considered that _ = 0:3. In (5), S is saliency map which is not binary. W and H are the width and height of the saliency map and x and y are number of each pixel. All the analytics data presented here are accepted as standard evaluations for saliency detection task but might not be meaningful and understandable for humans. Visual evaluation alongside with above analytics is the best way to judge network's performance. Precision and recall usually are not able to well describe the quality of saliency map. Similiary F-measure also favors saliency maps with high positive values.

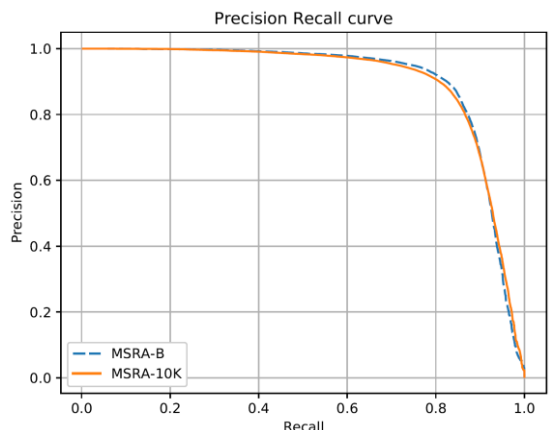

Fig. 2. Precision-Recall curve

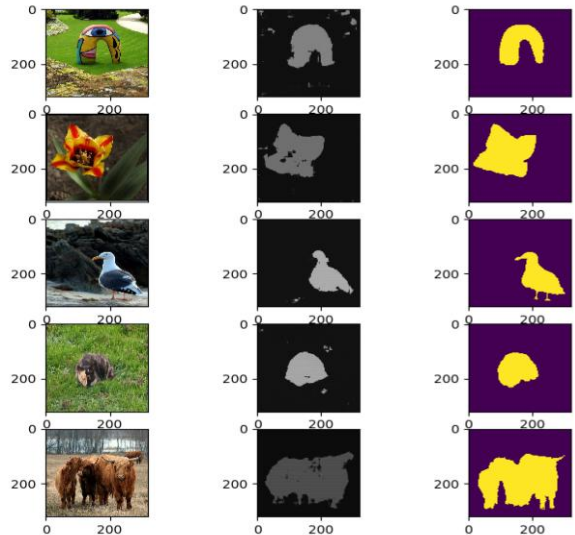

Fig. 3. (a) Original image (b) Saliency map from the network (c) Ground truth.

Mean absolute error is considered a better metric because it is applied on contineous map not binary, besidesit values the correctness of non-salient pixels too. The precision-recall curve for these two networks is shown in Fig. 2. The overall behavior of this curve shows the better performance of the network on MSRA-10k andMSRAB datasets, also other metrics of the network are illustrated in TABLE II. It can be seen that the network performed better on MSRA-B dataset which is smaller and is usually a standard for comparison, it had almost the same F-measure, it also was better around 1 percent on MEA metric. A few test examples from MSRA-B dataset along with detected saliency map and the corresponding ground truth is shown in Fig. 3.

These results show the importance of extracting different sized features for saliency detection task. The results between two datasets are close to eachother which shows the consistency of the network because MSRA-10k and MSRA-B images are very similar. There is also a comparison between the proposed network and two of the state of the art: PISA [15] and MDF [7]. Table III shows the performance of these methods. Table III shows that the presented mthod in this paper has better accuracy than other state of the art methods.

## CONCLUSION

In this paper a fully convolutional neural network was presented for saliency detection task. Since the network is fully convolutional this method is applicable on every input image size. With simple approach of using average poolingand max pooling and using batch normalization in up sampling layers, we were able to produce acceptable results on standard datasets. This network outperforms several other more complicated methods which used other parallel processes with CNN.

TABLE II: EVALUATION OF PRESENTED NETWORKS

| Test Dataset | F-measure | MAE | Precision | Recall |
|---|---|---|---|---|
| *MSRA-10K* | 0.785 | 0.102 | 0.796 | 0.795 |
| *MSRA-B* | 0.778 | 0.095 | 0.783 | 0.814 |

TABLE III: COMPARISON OF F-MEASURE AND MAE OF DIFFERENT METHODS

| Dataset | metric | PISA | MDF | ours |
|---|---|---|---|---|
| MSRA-B | AVG F-measure | 0.830 | 0.885 | 0.0874 |
| | MAE | 0.102 | 0.066 | 0.095 |

Besides by replacing maxpooling layers with average pooling we were able to pass more data from input to final layerswithout using shortcuts thus the network does less operations in last layers.

**REFERENCES**

[1] R. Achanta, F. Estrada, P. Wils, and S. S¨usstrunk, "Salient region detection and segmentation," in International conference on computer vision systems, pp. 66–75, Springer, 2008.

[2] P. F. Felzenszwalb and D. P. Huttenlocher, "Efficient graph-based image segmentation," International journal of computer vision, vol. 59, no. 2, pp. 167–181, 2004.

[3] D. Comaniciu and P. Meer, "Mean shift: A robust approach toward feature space analysis," IEEE Transactions on pattern analysis and machine intelligence, vol. 24, no. 5, pp. 603–619, 2002.

[4] D. M. Ram´ık, C. Sabourin, R. Moreno, and K. Madani, "A machine learning based intelligent vision system for autonomous object detection and recognition," Applied intelligence, vol. 40, no. 2, pp. 358–375, 2014.

[5] Y. Guo, Y. Liu, A. Oerlemans, S. Lao, S. Wu, and M. S. Lew, "Deep learning for visual understanding: A review," eurocomputing, vol. 187, pp. 27–48, 2016.

[6] G. Li and Y. Yu, "Deep contrast learning for salient object detection," in Proceedings of the IEEE Conference on Computer Vision and Pattern Recognition, pp. 478–487, 2016.

[7] G. Li and Y. Yu, "Visual saliency based on multiscale deep features," in Proceedings of the IEEE conference on computer vision and pattern recognition, pp. 5455–5463, 2015.

[8] H. Misaghi, R. A. Moghadam, and K. Madani, "Convolutional neural network for saliency detection in images," in Fuzzy and Intelligent Systems (CFIS), 2018 6th Iranian Joint Congress on, pp. 17–19, IEEE, 2018.

[9] R. Zhao, W. Ouyang, H. Li, and X. Wang, "Saliency detection by multi-context deep learning," in Proceedings of the IEEE Conference on Computer Vision and Pattern Recognition, pp. 1265–1274, 2015.

[10] K. Simonyan and A. Zisserman, "Very deep convolutional networks for large-scale image recognition," arXiv preprint arXiv:1409.1556, 2014.

[11] M.-M. Cheng, N. J. Mitra, X. Huang, P. H. Torr, and S.-M. Hu, "Global contrast based salient region detection," IEEE Transactions on Pattern Analysis and Machine Intelligence, vol. 37, no. 3, pp. 569–582, 2015.

[12] G. Li and Y. Yu, "Deep contrast learning for salient object detection," in Proceedings of the IEEE Conference on Computer Vision and Pattern Recognition, pp. 478–487, 2016.

[13] D. P. Kingma and J. Ba, "Adam: A method for stochastic optimization," arXiv preprint arXiv:1412.6980, 2014.

[14] Q. Hou, M.-M. Cheng, X. Hu, A. Borji, Z. Tu, and P. Torr, "Deeply supervised salient object detection with short connections," IEEE TPAMI, 2018.

[15] K. Wang, L. Lin, J. Lu, C. Li, and K. Shi, "Pisa: Pixelwise image saliency by aggregating complementary appearance contrast measures with edge-preserving coherence," IEEE Transactions on Image Processing, vol. 24, no. 10, pp. 3019–3033, 2015.

★★★